\pdfoutput=1

\documentclass[11pt]{article}

\usepackage[]{acl}

\usepackage{times}
\usepackage{latexsym}

\usepackage[T1]{fontenc}

\usepackage[utf8]{inputenc}

\usepackage{microtype}

\usepackage{inconsolata}

\usepackage{graphicx}

\usepackage{amsmath}
\usepackage{booktabs}
\usepackage{pgfplots}
\usepackage{pgfplotstable}
\usepackage{graphicx}
\usepackage{multirow}
\usepackage{amsmath}
\usepackage{adjustbox}
\usepackage{float}
\usepackage{ulem}
\usepackage{tabularx}
\usepackage{array}
\usepackage{amsfonts}

\title{QUPID: Quantified Understanding for Enhanced Performance, Insights, and Decisions in Korean Search Engines}

\author{
    \textbf{Ohjoon Kwon\thanks{Equal contribution}},
    \textbf{Changsu Lee\footnotemark[1]},
    \textbf{Jihye Back}, 
    \textbf{Lim Sun Suk}, \\
    \textbf{Inho Kang},
    \textbf{Donghyeon Jeon\thanks{Corresponding author}} \\
    Naver Corporation \\
    \texttt{\{ohjoon.kwon, changsu.lee, 1oojihye, dongle.75}, \\
    \texttt{once.ihkang, donghyeon.jeon\}@navercorp.com}
}

\begin{document}
\maketitle
\begin{abstract}

Large language models (LLMs) have been widely used for relevance assessment in information retrieval. However, our study demonstrates that combining two distinct small language models (SLMs) with different architectures can outperform LLMs in this task. Our approach—QUPID—integrates a generative SLM with an embedding-based SLM, achieving higher relevance judgment accuracy while reducing computational costs compared to state-of-the-art LLM solutions. This computational efficiency makes QUPID highly scalable for real-world search systems processing millions of queries daily. In experiments across diverse document types, our method demonstrated consistent performance improvements (Cohen's Kappa of 0.646 versus 0.387 for leading LLMs) while offering 60x faster inference times. Furthermore, when integrated into production search pipelines, QUPID improved nDCG@5 scores by 1.9\%. These findings underscore how architectural diversity in model combinations can significantly enhance both search relevance and operational efficiency in information retrieval systems.
\end{abstract}

\section{Introduction}

Large Language Models (LLMs) have demonstrated remarkable capabilities in Information Retrieval (IR) tasks, including query-document relevance assessment \citep{li2024matchinggenerationsurveygenerative, zhu2024largelanguagemodelsinformation, tang2024selfretrievalendtoendinformationretrieval}. Their strong contextual understanding allows them to approximate human-level judgments in ranking search results and evaluating query modifications. However, deploying LLMs in production environments comes with significant challenges. Latency issues pose the most critical barrier for large-scale search systems, where real-time responses are essential, followed by the high computational cost and substantial memory footprint, making them impractical for environments operating on a scale of hundreds of millions of queries daily \citep{strubell2019energypolicyconsiderationsdeep, sharir2020costtrainingnlpmodels, howell2023economictradeoffslargelanguage, thomas2024largelanguagemodelsacc}. Furthermore, maintaining and continuously updating such models is resource-intensive, limiting their feasibility in dynamic and multilingual search environments.

To address these challenges, Small Language Models (SLMs) have emerged as a promising alternative. SLMs offer a more cost-effective solution while maintaining competitive performance in various NLP tasks \citep{10.1145/3616855.3635690, brei2024leveragingsmalllanguagemodels, xu2025slmrecdistillinglargelanguage}. However, existing approaches to SLM-based relevance assessment face two critical limitations: (1) Single SLM approaches lack the expressive power and contextual understanding of LLMs, resulting in suboptimal relevance judgments in complex queries; and (2) Current ensemble methods primarily utilize multiple instances of the same model architecture \citep{rahmani2024judgeblenderensemblingjudgmentsautomatic}, failing to leverage the complementary strengths that different model architectures could provide. 

In this study, we propose QUPID (Quantified Understanding for Enhanced Performance, Insights, and Decisions), a novel relevance assessment approach that directly addresses these limitations by combining two architecturally distinct SLMs in a heterogeneous ensemble. Unlike prior ensemble methods that aggregate multiple instances of the same model, our approach integrates a generative SLM ($\text{QUPID}_{GEN}$), which excels at contextual reasoning through token probabilities, with an embedding-based SLM ($\text{QUPID}_{EMB}$), which captures semantic similarity through dense vector representations. By leveraging both generative reasoning and embedding-based similarity within a unified framework, QUPID outperforms state-of-the-art LLMs and SLM ensemble baselines while significantly reducing computational cost. Our results show that this heterogeneous ensemble strategy enhances relevance labeling accuracy, making it a scalable and efficient solution for real-world search engines.



Beyond accuracy improvements, our model has broad practical implications in modern search systems. We demonstrate how QUPID can be integrated into various workflows, including filtering low-quality query-document pairs, evaluating query rewriting and completion modules, assessing the quality of search result snippets, and enhancing ranking models.

To summarize, our contributions are as follows:

\begin{itemize}
\item We introduce QUPID, the first relevance labeling approach leveraging a heterogeneous ensemble of generative and embedding-based SLMs, highlighting the effectiveness of architectural diversity.
\item We demonstrate that this approach outperforms state-of-the-art LLM-based methods by up to 67\% in terms of Cohen's Kappa ($\kappa$ = 0.646 vs. $\kappa$ = 0.387), while offering 60x faster inference times (62ms vs. 3258ms), making it highly scalable for production environments.
\item We present concrete use cases showcasing how QUPID can be seamlessly integrated into search engine pipelines, yielding measurable improvements in retrieval quality metrics (e.g., +1.9\% in nDCG@5) and enhancing overall user satisfaction in real-world deployments.
\end{itemize}

\section{Related Works}
\paragraph{LLM-Based Relevance Labeling}
Recent studies have explored the potential of Large Language Models (LLMs) in relevance assessment tasks, leveraging their strong contextual reasoning and generation capabilities \citep{li2024matchinggenerationsurveygenerative, thomas2024largelanguagemodelsacc, upadhyay2024umbrelaumbrelaopensourcereproduction}. For example, \citet{thomas2024largelanguagemodelsacc} demonstrated that LLMs could achieve near-human performance in evaluating query-document relevance. However, their proprietary, in-house model setup makes exact replication difficult, limiting reproducibility in real-world applications. Similarly, \citet{upadhyay2024umbrelaumbrelaopensourcereproduction} introduced an open-source toolkit for evaluating LLM-based relevance labeling models, but their results revealed significant computational inefficiencies, especially when applied to high-throughput search systems. Moreover, these LLM-based methods generally perform well in English but show suboptimal results in multilingual environments such as Korean, necessitating lighter, language-specific models for more efficient deployment \citep{robinson-etal-2023-chatgpt, bang2023multitaskmultilingualmultimodalevaluation, nguyen2024democratizingllmslowresourcelanguages, li2024languagerankermetricquantifying, jayakody2024performancerecentlargelanguage}.

\paragraph{Embedding-Based Relevance Labeling in LLMs}
Beyond text generation, recent work has shown that decoder-only LLMs can also be used for embedding-based retrieval tasks \citep{ma2023finetuningllamamultistagetext, wang2024improvingtextembeddingslarge, li-etal-2024-llama2vec}. These models, originally trained for next-token prediction, exhibit surprising representation learning capabilities that can be leveraged for similarity-based tasks. However, most existing studies in this domain have focused on fine-tuning embedding models for generalized text similarity, rather than specifically optimizing for query-document relevance labeling.

To address these limitations, several approaches have attempted to enhance LLM embeddings by modifying the attention mechanisms or introducing hard-negative mining techniques \citep{wang2024improvingtextembeddingslarge, li-etal-2024-llama2vec}. Building on this line of work, we integrate embedding capabilities into our SLM ensemble, making it the first hybrid generative + embedding-based SLM model for relevance assessment. This allows our approach to leverage both explicit generation-based judgments and implicit similarity-based signals, improving robustness and accuracy. By uniting explicit generation-based judgments (which capture contextual cues and semantic coherence) with implicit similarity-based signals (which highlight distributional proximity between query and document), our model can more accurately reflect both high-level language understanding and fine-grained relational cues. 

\paragraph{SLM-Based Ensembles for Relevance Assessment}
To address the high computational costs of LLMs, recent studies have explored ensemble methods using Small Language Models (SLMs). For example, JudgeBlender \citep{rahmani2024judgeblenderensemblingjudgmentsautomatic} aggregates multiple generative SLMs to improve relevance labeling accuracy. However, this approach requires long prompts and a two-step inference process, leading to higher inference latency (1000ms+) and resource consumption. Furthermore, it focuses solely on generation-based relevance assessment without incorporating embedding-based similarity, potentially limiting its effectiveness in ranking tasks. This synergy yields richer feature representations and mitigates the limitations of using either approach alone, ultimately enhancing both robustness and accuracy in query-document relevance tasks.

Unlike these prior works, our approach introduces a heterogeneous ensemble of two distinct SLM architectures, where one model specializes in generative reasoning and the other in embedding-based relevance computation. By combining both generative and representation-learning capabilities, our method achieves higher accuracy with significantly reduced computational cost, making it scalable for real-world search systems.

\section{Methodology}\label{section:Methodology}
In this section, we introduce our approach to relevance labeling for query-document pairs using two small-scale language models (SLMs). We first describe the process of creating and cleaning our dataset (Section \ref{subsec:data curation}). We then employ one SLM as a generative relevance labeling model and another SLM as an embedding-based relevance labeling model (detailed in Sections \ref{subsec:gen model} and \ref{subsec:emb model}, respectively). Finally, we describe our ensemble strategy that combines both models’ outputs to achieve higher accuracy and robustness (Section \ref{subsec:score ensemble}).

\subsection{Data Curation}\label{subsec:data curation}
Our approach to curating training data for relevance labeling consists of two main strategies: collecting real-world document data and generating synthetic hard-negative samples. By integrating these approaches, we ensure that our dataset is diverse, well-balanced, and robust to generalization challenges.

\subsubsection{Real-World Document Collection} \label{subsubsection: real data}
We collected various query-document pairs from the web. These pairs come from three main sources to capture a broad range of text lengths, styles, and complexities:

\begin{itemize}
\item Snippet-Style Web Content: Short pieces of text—often lists, tables, or brief paragraphs—extracted from search engine snippets.
\item User-Generated Content: Texts directly provided or created by users, such as forum posts or community Q\&A entries.
\item General Web Documents: Freely crawled online documents covering diverse topics and domains.
\end{itemize}

With more than 850K query-text pairs, experienced human annotators\footnote{The annotators are part of a specialized data-labeling company and work under close guidance and feedback from linguists with domain expertise.} carefully review each pair and assign an appropriate label among four classes (see Appendix~\ref{appendix: guidelines for evaluating} for details). We adopt a voting scheme to resolve any disagreements, and if all three annotators assign different labels, an additional linguist overseeing the annotation process makes the final judgment. By including snippet-style, user-generated, and general web documents, we ensure varied document lengths and structures such as text, lists, and tables, which is vital for robust model training. (Note that \textit{Somewhat Relevant} has been classified as a relevant label.)

\subsubsection{Synthetic Hard-Negatives Generation} \label{subsubsection: synthetic hard negatives}
Existing studies in retrieval and embedding emphasize the importance of hard-negative samples to enhance model robustness and generalization \citep{lee2024geckoversatiletextembeddings, moreira2024nvretrieverimprovingtextembedding, Aho:72, wang2024improvingtextembeddingslarge}. We follow a similar strategy to further refine our training data.

For each query-document pair collected in Section \ref{subsubsection: real data}, we prompt the model to generate new documents that are superficially similar to the original but contextually off-topic or partially misleading. We use Mistral2-Large to produce additional hard-negative documents because it shows plausible Korean synthetic data based on our internal evaluation. Further details on the prompt design can be found in Appendix \ref{subsection: hard-negative generation}.

By combining real-world data (approximately 48.70\% web documents, 12.17\% user-generated content, and 24.62\% snippets) with synthetically generated hard negatives (14.51\%), we construct a curated dataset comprising roughly 1M query-document pairs.


\subsection{Generative Relevance Labeling Model}\label{subsec:gen model}
\paragraph{SLM as a Relevance Labeler}
When using a generative model for relevance labeling in a query-document setting, there are generally two approaches. One is to directly interpret the tokens produced by the large language model (LLM) as the final decision, optionally appending a confidence estimation step \citep{thomas2024largelanguagemodelsacc, ni2025dirasefficientllmannotation}. Another approach is to leverage the token probabilities associated with a specific label or query \citep{sachan2023improvingpassageretrievalzeroshot, zhuang2024yesnoimprovingzeroshot}, thereby introducing a more explicit calibration stage.

In this work, we follow the token-probability approach proposed by \citet{zhuang2024yesnoimprovingzeroshot}, rather than interpreting directly sampled tokens. Specifically, for each query-document pair 
\((q_i, d_i)\), our generative model 
\(M_{\mathrm{gen}}\) is fine-tuned to produce exactly one among three special tokens 
\(\{l_1, l_2, l_3\}\). Each token corresponds to a distinct relevance category (e.g., Highly Relevant, Low Relevance, Not Relevant). We then obtain the log probability of each label token and convert these values into probabilities via a softmax function as follows:
\begin{equation}
p_{i,k} = \mathrm{M_{gen}}(l_k \mid q_i, d_i),
\label{eq:prob_llm}
\end{equation}
be the probability that the model assigns to label \(l_k\). We define the final relevance score as:
\begin{equation}
f(q_i, d_i) 
= \sum_{k=1}^{K} p_{i,k} \cdot y_k,
\label{eq:final_score}
\end{equation}
where \(y_k\) is a label-specific weight (or numeric value) that can be tuned based on downstream requirements or validation set performance. In our experiments, we set these values as follows: \(\textit{relevant}=1.0\), \(\textit{somewhat relevant}=0.5\), and \(\textit{irrelevant}=0\). This approach provides a more informative representation of the model's confidence by considering the probability distribution over possible labels rather than relying on a single sampled output. This allows for a more nuanced understanding of the model’s uncertainty, which can be particularly useful in downstream applications requiring risk-aware decision-making.

\subsection{Embedding-based Relevance Labeling Model}\label{subsec:emb model}
For fine-tuning the embedding-based model, \(M_{\mathrm{emb}}\), it takes a query-document pair \((q, d)\) as input and generates a relevance score through an embedding extraction and classification process. Given an input pair, the model produces a sequence of token-level hidden state embeddings:
\begin{equation}
H = M_{\mathrm{emb}}(q, d) = [h_1, h_2, \dots, h_n],
\label{eq:hidden_states}
\end{equation}

where \(h_i \in \mathbb{R}^{d_h}\) represents the hidden state of the \(i\)-th token, and \(d_h\) denotes the dimensionality of the model’s hidden representation. To obtain a fixed-size representation, we apply mean pooling as it showed more stable performance in our experiments and prior research \citep{lee2025nvembedimprovedtechniquestraining, moreira2025nvretrieverimprovingtextembedding}:
\begin{equation}
\mathbf{h}_{\text{agg}} = \frac{1}{n} \sum_{i=1}^{n} h_i.
\label{eq:mean_pooling}
\end{equation}
A linear transformation is then applied to map \(\mathbf{h}_{\text{agg}}\) into a relevance score vector:
\begin{equation}
s = W \mathbf{h}_{\text{agg}} + b,
\label{eq:linear_projection}
\end{equation}
where \(W \in \mathbb{R}^{K \times d_h}\) is the learned weight matrix, \(b \in \mathbb{R}^K\) is the bias term, and \(K\) represents the number of predefined relevance categories. Finally, the softmax function is applied to compute the probability distribution over the relevance labels.


\subsection{Score Ensemble}\label{subsec:score ensemble}
To enhance the robustness of relevance scoring, we combine the outputs of the generative model \(M_{\mathrm{gen}}\) and the embedding-based model \(M_{\mathrm{emb}}\) through a weighted averaging approach. Weighted averaging preserves the independence of each model’s outputs and provides a straightforward way to balance generative vs. embedding signals Each model produces a relevance score given a query-document pair \((q, d)\):
\begin{equation}
s_{\mathrm{gen}} = M_{\mathrm{gen}}(q, d),
\label{eq:gen_score}
\end{equation}
\begin{equation}
s_{\mathrm{emb}} = M_{\mathrm{emb}}(q, d).
\label{eq:emb_score}
\end{equation}
To obtain the final ensemble relevance score, we compute a weighted sum of these two scores:
\begin{equation}
s_{\mathrm{final}} = w_{\mathrm{gen}} \cdot s_{\mathrm{gen}} + w_{\mathrm{emb}} \cdot s_{\mathrm{emb}},
\label{eq:weighted_sum}
\end{equation}
where \(w_{\mathrm{gen}}\) and \(w_{\mathrm{emb}}\) are the weighting coefficients assigned to the generative and embedding-based models, respectively. The weights can be tuned based on validation performance or set heuristically. In practice, we determine the optimal \(w_{\mathrm{gen}}\) and \(w_{\mathrm{emb}}\) by evaluating the ensemble’s effectiveness on a held-out validation set, selecting the combination that maximizes relevance prediction accuracy.

\section{Experimental Setup}
In this section, we present a comprehensive evaluation environment of our model in diverse experimental settings. Our model is fine-tuned on HCX-S \citep{yoo2024hyperclovaxtechnicalreport}, a state-of-the-art instruct-tuned model known for its superior performance in Korean-language tasks.

\subsection{Dataset}
To evaluate robustness across document types, we build test sets comprising snippet-style web content (3,000 samples; 2,268 relevant, 732 irrelevant), user-generated content (20,000 samples; 11,148 relevant, 8,852 irrelevant), and general web documents (9,000 samples; 4,971 relevant, 4,029 irrelevant).

\begin{table*}[htbp]
    \centering
    \begin{adjustbox}{width=\textwidth}
        \begin{tabular}{l|cccc|cccc}
            \toprule
            \multirow{2}{*}{\textbf{Model}} & \multicolumn{4}{c|}{\textbf{Cohen’s Kappa ($\kappa$)}} & \multicolumn{4}{c}{\textbf{AUC (Relevant / Irrelevant)}} \\
            \cmidrule(lr){2-5} \cmidrule(lr){6-9}
            & \textbf{UGC} & \textbf{Snippet} & \textbf{Web-D} & Avg. & \textbf{UGC} & \textbf{Snippet} & \textbf{Web-D} & Avg. \\
            \midrule
            \multicolumn{9}{c}{\textbf{Representative LLMs}} \\
            ChatGPT-4o & 0.224 & 0.424 & 0.514 & 0.387 & 0.696 / 0.622  & 0.915 / 0.511 & 0.818 / 0.760 & 0.810 / 0.631  \\
            LLaMA3.3-70b-instruct & 0.129 & 0.256 & 0.402 & 0.262 & 0.667 / 0.141 & 0.883 / 0.229 & 0.762 / 0.533 & 0.771 / 0.301 \\
            Mistral-large-instruct-2411 & 0.154 & 0.326 & 0.458 & 0.313 & 0.713 / 0.632 & 0.924 / 0.578 & 0.812 / 0.751 & 0.816 / 0.654 \\
            Qwen-2.5-72b-instruct & 0.160 & 0.333 & 0.436 & 0.310 & 0.721 / 0.654 & 0.954 / 0.574 & 0.821 / 0.762 & 0.832 / 0.663 \\        
            \midrule
            \multicolumn{9}{c}{\textbf{SLM Ensemble Model}} \\
            Mistral-8b-instruct-2410 & 0.151 & 0.119  & 0.257 & 0.176 & 0.695 / 0.542 & 0.781 / 0.363 & 0.634 / 0.494 & 0.703 /0.466 \\
            Qwen-2.5-7b-instruct & 0.124 & 0.262 & 0.318 & 0.235 & 0.698 / 0.379 & 0.839 / 0.419 & 0.711 / 0.564 & 0.749 / 0.454 \\
            HCX-S & 0.157 & 0.174 & 0.298 & 0.210 & 0.692 / 0.547 & 0.783 / 0.415 & 0.645 / 0.564 & 0.707 / 0.509 \\
            JudgeBlender &  &  &  &  &  &  &  &  \\
            +MV(Avg.) & 0.207 & 0.269 & 0.291 & 0.256 & 0.696 / 0.640 & 0.867 / 0.472 & 0.686 / 0.710  & 0.750 / 0.607 \\
            +MV(Rand.) & 0.162 & 0.183 & 0.298 & 0.214 & 0.633 / 0.482 & 0.783 / 0.396 & 0.595 / 0.633 & 0.670 / 0.504 \\
            +AV & 0.154 & 0.278 & 0.331 & 0.254 & 0.652 / 0.623 & 0.881 / 0.494 & 0.689 / 0.724  & 0.741 / 0.614 \\
            \midrule
            \multicolumn{9}{c}{\textbf{Ours (fine-tuned)}} \\
            $\text{QUPID}_{GEN}$ & 0.582 & 0.418 & 0.674 & 0.558 & 0.897 / 0.880 & 0.932 / 0.707 & 0.960 / 0.940 & 0.930 / 0.842 \\
            $\text{QUPID}_{EMB}$ & 0.662 & 0.518 & 0.590 & 0.590 & 0.927 / 0.892 & 0.904 / 0.715 & 0.889 / 0.851 & 0.907 / 0.819 \\
            $\text{QUPID}_{ENSEMBLE}$ & \textbf{0.679} & \textbf{0.569} & \textbf{0.783} & \textbf{0.646} & \textbf{0.929} / \textbf{0.911} & \textbf{0.944} / \textbf{0.756} & \textbf{0.962} / \textbf{0.946} & \textbf{0.945} / \textbf{0.871}  \\
            \bottomrule
        \end{tabular}
    \end{adjustbox}
    \caption{Evaluation results of different models on three datasets. Cohen’s Kappa ($\kappa$) and AUC scores are reported. AUC is reported as Relevant AUC / Irrelevant AUC.}
    \label{tab:main_evaluation}
\end{table*}

\subsection{Baselines}
To evaluate the effectiveness of our proposed hybrid ensemble approach, we compare our models against two categories of baselines: (1) representative large language models (LLMs) and (2) SLM ensemble models.

\subsubsection{Representative LLMs}
We evaluate a set of representative LLMs (LLaMA, Mistral, and Qwen), each capable of performing relevance labeling via zero-shot inference. Various prompting strategies exist for instructing LLMs in relevance assessment \citep{sun-etal-2023-chatgpt, Faggioli_2023, farzi2024besttaullmjudgecriteriabasedrelevance, thomas2024largelanguagemodelsacc}, and we experiment with the representative prompting method showed best performance in LLMJudge benchmark \citep{rahmani2024llmjudgellmsrelevancejudgments}. For more information, see the Table 3 in \citet{farzi2024besttaullmjudgecriteriabasedrelevance}

\subsubsection{SLM Ensemble Method}
We also compare our approach to JudgeBlender \citep{rahmani2024judgeblenderensemblingjudgmentsautomatic}. It follows a multi-model ensembling strategy where each constituent model is prompted separately for relevance assessment, and their outputs are combined to improve robustness. The same prompts used in the representative LLMs experiments were employed. Based on the observation that LLaMA models performs poorly in Korean, we replaced it with the HCX-S \citep{yoo2024hyperclovaxtechnicalreport}.

\section{Results}
\subsection{Quantitative Results}
Table \ref{tab:main_evaluation} reveals that representative large language models (LLMs) face challenges in capturing the precise relevance between queries and documents. LLaMA3.3-70b shows notably lower performance, likely due to its weaker multilingual capabilities, especially in Korean. To provide a comprehensive evaluation of model effectiveness, we employed AUC to assess how well each model ranks relevant documents against irrelevant ones across varying thresholds. Additionally, we used Cohen's kappa to measure the agreement level between model predictions and human-annotated relevance labels, offering insights into how closely automated labeling aligns with human judgment.

While the ensemble methods with SLMs show some improvement, they are still not enough to replace human judgment or reliably assess search quality. The two-step inference and ensemble approach based on three SLMs of approximately 8B parameters fell short of the zero-shot prompting performance of the LLMs. This suggests that a simple SLM ensemble, without target-specific fine-tuning or heterogeneous model combinations, cannot match the capacity of LLMs.

Our proposed model consistently outperforms these leading LLMs and SLM blending methods. Moreover, Table \ref{tab:heterogeneous_results} demonstrates that the combination of heterogeneous models in an ensemble leads to substantial performance improvements, highlighting the advantage of leveraging diverse model architectures to enhance task effectiveness.

\begin{table}[ht]
    \centering
    \begin{tabular}{lrr}
        \toprule
        \textbf{Model} & \textbf{Avg. $\kappa$} & \textbf{Avg. AUC} \\
        \midrule
        $\text{QUPID}_{GEN}$ & 0.558 & 0.930 / 0.842 \\
        $\text{QUPID}_{GEN*3}$ & 0.564 & 0.932 / 0.849 \\
        $\text{QUPID}_{GEN*5}$ & 0.569 & 0.933 / 0.851 \\
        \midrule
        $\text{QUPID}_{EMB}$ & 0.590 & 0.907 / 0.819 \\
        $\text{QUPID}_{EMB*3}$ & 0.597 & 0.913 / 0.832 \\
        $\text{QUPID}_{EMB*5}$ & 0.607 & 0.913 / 0.835 \\        
        \midrule
        $\text{QUPID}_{ENSEMBLE}$ & \textbf{0.646} & \textbf{0.945 / 0.871} \\        
        \bottomrule
    \end{tabular}
    \caption{The rows above show the results of ensembling with the same architecture that were trained using the same approach but with different hyperparameters.}
    \label{tab:heterogeneous_results}
\end{table}

\subsection{Use Cases and Efficiency}
We examine several practical use cases demonstrating how our QUPID can be seamlessly integrated into real-world search engine workflows. These particular use cases—(1) filtering low-quality pairs, (2) evaluating query rewriting and completion, (3) assessing snippet quality, and (4) improving ranking—were selected because they represent common challenges in large-scale search systems and highlight different facets of relevance assessment.

\paragraph{Filtering low-quality Q-D pairs}
Search engines pre-assign documents to frequently occurring or time-sensitive queries to enable faster response times \citep{nogueira2019documentexpansionqueryprediction, 10.1145/3539618.3592028}. Since these documents often appear at the top of search results with high confidence, ensuring their relevance and quality is crucial. As shown in Appendix~\ref{subsub: filtering Q-D pairs} and Figure~\ref{fig:pr-curve}, QUPID plays a vital role in identifying and filtering out low-quality content before it reaches users, achieving a precision of over 0.9.

\paragraph{Evaluating Query Refinement Modules}
Our relevance model provides an automatic approach to evaluating query rewriting \citep{wu2022conqrrconversationalqueryrewriting, sun2024rbotllmbasedqueryrewrite, liu2024queryrewritinglargelanguage} or completion models \citep{jaech-ostendorf-2018-personalized, kim-2019-subword, Gog_2020}. Effectiveness of these models should be assessed based on whether they improve the search results. As illustrated in Table \ref{tab:qr_qc_comparison}, we utilize our relevance model to evaluate the performance of the LLM-based query generation modules operating in our search engine.

\paragraph{Evaluating the quality of snippets extracted from documents}
Our relevance model can independently evaluate query-document (Q-D) relevance and query-snippet (Q-S) relevance. If a document has a high relevance score but a low snippet relevance score, it suggests that the document itself is relevant to the query, but the extracted snippet is misleading or unrepresentative. As illustrated in Table \ref{tab:side-by-side-comparison of document and snippet}, such cases often lead to inaccurate search summaries, which can negatively impact the user experience. 

\paragraph{Applying QUPID for Search Results Ranking}
To further evaluate the benefits of QUPID in a real-world ranking scenario, we tested the model on the ranking task. Table~\ref{tab:ranking_performance} shows the ranking metrics achieved by the baseline ranking model and our proposed method. Having explored the potential of replacing the existing ranking model, we actively utilize QUPID as the ranking model in our search engine.

\paragraph{Efficiency Compare}
For efficiency evaluation, we measured the latency of each model. Each model was deployed and served on the same A100-80G GPUs with vLLM serving engine. Due to a significantly shorter system prompt and generating at most a single token, the QUPID model exhibits overwhelmingly faster latency.


\begin{table}[ht]
    \centering
    \begin{tabular}{p{3.1cm} p{2.0cm} p{1.27cm}} 
        \toprule
        \textbf{Model} & \textbf{Sys. Prompt} & \textbf{Latency} \\
        \midrule
        $\text{QUPID}_{ENSEMBLE}$ & 10 token & 62 ms  \\
        \midrule
        JudgeBlender & 362 tokens & 1173 ms \\
        \midrule
        LLaMA3.3-70b & 353 tokens & 3258 ms  \\
        Qwen-2.5-72b & 384 tokens & 3520 ms  \\
        Mistral-large & 392 tokens & 3690 ms \\
        \bottomrule
    \end{tabular}
    \caption{The input tokens for JudgeBlender are the average of the three models. When results from multiple models are required, they are obtained asynchronously through parallel calls. Refer to Appendix \ref{appendix:efficiency compare details}.}
    \label{tab:efficiency compare table}
\end{table}

\section{Conclusion}
In this paper, we introduced QUPID, a heterogeneous ensemble of generative and embedding-based SLMs for relevance labeling. Our approach improves accuracy while reducing computational costs and demonstrates practical applicability in real-world search engines. Our findings highlight the value of model diversity in enhancing relevance assessment while maintaining efficiency. 

\section{Limitations}
While our proposed QUPID approach demonstrates significant improvements in relevance labeling efficiency and accuracy, it has several limitations that warrant further investigation.

\paragraph{Text Modality Only} 
Our current approach is designed exclusively for text-based relevance assessment and does not incorporate multimodal capabilities. Many modern search applications involve a combination of text, images, and videos, where relevance cannot always be determined through textual information alone. The lack of image and video processing limits the applicability of our method in broader search engine contexts, particularly in domains such as e-commerce, news aggregation, and multimedia search.

\paragraph{Limited Feature Scope} 
Our model primarily focuses on relevance as the key criterion for assessing search results. However, in real-world applications, additional factors such as readability, aesthetic appeal, and trustworthiness also play a critical role in determining the overall quality of retrieved documents. For instance, a document may be highly relevant but poorly formatted or difficult to read, negatively impacting user experience. Future iterations of our model should incorporate auxiliary scoring mechanisms to address these qualitative aspects of search quality.

Despite these limitations, our findings establish a strong foundation for efficient and scalable relevance assessment. Addressing these challenges in future research—particularly through multimodal expansion and the inclusion of richer feature representations—will further enhance the practicality and robustness of our approach.

\section*{Acknowledgments}
We would like to express our heartfelt gratitude to the labeling experts for their invaluable assistance in reviewing and verifying the solutions presented in this work. Their detailed feedback, as well as the efforts of all other contributors, significantly improved the quality and accuracy of our results. We confirm that each contributor was duly compensated for their time and expertise. Additionally, we utilized generative AI tools to assist with code development and to refine the English language in this manuscript. The authors assume full responsibility for the content herein, including any errors that may remain.

\bibliography{custom}

\appendix

\section{Appendices}
 
\subsection{Guidelines for Evaluating Search Results} \label{appendix: guidelines for evaluating}

\subsubsection{Evaluation Scope}
The evaluation targets documents (text) corresponding to a query. Both the title and body content are considered holistically. The key criterion is whether the title and body together provide sufficient and relevant information to match the user’s intent.

\subsubsection{Evaluation Criteria}
Documents are classified into four categories:
\begin{itemize}
    \item \textbf{Relevant (R)}: Fully relevant and contains sufficient information.
    \item \textbf{Somewhat Relevant (SR)}: Relevant but lacks sufficient details.
    \item \textbf{Irrelevant (I)}: Either irrelevant or missing key information.
    \item \textbf{Not Evaluable}: Cases where the query is unclear or improperly corrected by a search suggestion, making them unsuitable for reliable annotation. As such, they are excluded from both training and evaluation.

\end{itemize}

\subsubsection{Key Evaluation Considerations}
\begin{itemize}
    \item \textbf{Title \& Body Alignment}: A document is considered high quality if the title and body together sufficiently answer the query.
    \item \textbf{Insufficient Body Content}: Even if the title is relevant, a document is marked as low quality if the body lacks necessary details.
    \item \textbf{Title-Body Discrepancy}: If the title does not match the query but the body contains sufficient information, the document is still rated based on body content.
    \item \textbf{Hashtag-Based Content}: Hashtags alone can be evaluated if they effectively convey information.
    \item \textbf{Non-Text Queries}: Queries containing only numbers or foreign languages are excluded.
\end{itemize}

\subsubsection{Evaluation Process}
\begin{enumerate}
    \item \textbf{Relevance Check}: Determines if the document aligns with the user’s intent.
    \begin{itemize}
        \item If unrelated, it is marked \textbf{Irrelevant (I)}.
        \item If related, move to the next step.
    \end{itemize}
    \item \textbf{Information Sufficiency Check}:
    \begin{itemize}
        \item If the document fully answers the query, it is \textbf{Relevant (R)}.
        \item If additional searches are needed, it is \textbf{Somewhat Relevant (SR)}.
        \item If the document lacks necessary details entirely, it is \textbf{Irrelevant (I)}.
    \end{itemize}
    \item \textbf{Freshness Requirement}: If the query demands the latest data (e.g., financial, legal, or event-based queries), outdated responses are marked \textbf{Irrelevant (I)}.
\end{enumerate}

\subsubsection{Handling of Special Cases}
\begin{itemize}
    \item \textbf{Ambiguous Queries}: If a query has multiple meanings, the most commonly searched intent is used for evaluation.
    \item \textbf{Search Correction Errors}: If an automatic query correction leads to a mismatched query-document pair, the result is marked \textbf{Not Evaluable}.
    \item \textbf{Table/List Format Documents}: If extracted tables contain errors, missing data, or require verification, they are marked for \textbf{Further Review}.
\end{itemize}

This structured framework ensures objective and consistent evaluation of search results.

\subsection{Hard Negative Document Generation}\label{subsection: hard-negative generation}

To enhance the robustness of our ranking model, we generate hard negative documents by leveraging a structured prompt. The objective is to create documents that contain query-related keywords but deviate in meaning, ensuring they do not fulfill the user's intent. This method helps the model distinguish between relevant and misleading results.

\subsubsection{Structured Prompt for Hard Negative Document Generation}
We utilize the following structured prompt to systematically generate hard negative documents. This prompt ensures that the generated documents resemble real-world documents while maintaining low relevance to the given query.

\begin{quote}
\textbf{System Prompt:} You are an AI search system optimized for retrieving relevant information based on user queries.

\textbf{Instruction:} Given a \textbf{search query} and its \textbf{highest relevance document}, generate a new hard negative document that meets the following criteria:

\begin{itemize}
    \item The document must contain keywords from the query but use them in a different semantic context.
    \item The document should provide useful information, but the information must not align with the query’s intent.
    \item The document should not contain any direct answers to the query but may include peripheral information.
    \item The document must be factually accurate and must not include fabricated or false information.
    \item The document should be significantly less relevant to the query compared to the most relevant document.
\end{itemize}

Additionally, the document’s style should match the style of the given relevant document:
\begin{itemize}
    \item If the most relevant document follows an encyclopedic format, the generated document should also adopt a formal, academic style.
    \item If the most relevant document follows a blog-like format, the generated document should adopt a conversational and subjective style.
\end{itemize}

Ensure that the generated output follows these guidelines and is formatted in JSON with a clear distinction between \texttt{title} and \texttt{document} fields.
\end{quote}

\subsubsection{Criteria for Hard Negative Documents}
Hard negative documents are generated based on the following criteria:

\begin{itemize}
    \item \textbf{Query Mismatch:} The document discusses a topic that is lexically similar to the query but semantically different.
    \item \textbf{Useful but Irrelevant:} The document contains valuable information but does not directly answer the query.
    \item \textbf{Incomplete Information:} The document provides only partial information, requiring further searches to obtain the full answer.
\end{itemize}

\subsection{Hyper Parameters}\label{sec:appendix-hyperparams}
In this section, we detail the hyperparameters used for training
the embedding-based model (\textbf{EMB}) and the generative model
(\textbf{GEN}) described in our methodology. We also indicate
the proportion of synthetic data used, along with any notable
implementation remarks. Refer to Table \ref{tab:hyperparams}.

\begin{table}[t]
\centering
\begin{tabular}{p{2.2cm}p{2.2cm}p{2.2cm}}  
\toprule
\textbf{Hyperparams.} & $\text{QUPID}_{EMB}$ & $\text{QUPID}_{GEN}$ \\
\midrule
\textbf{LR} & 1e-5 & 1e-5 \\
\textbf{Epochs}        & 5    & 5 \\
\textbf{Optimizer} & Adam & Adam \\
\textbf{GPU (hours)} & A100 (440) & A100 (480) \\
\textbf{Max Length} & 1024 & 1024 \\
\bottomrule
\end{tabular}
\caption{T: inference temperature. Both model trained on 8xA100-80G.}
\label{tab:hyperparams}
\end{table}

\begin{figure}[t] 
    \centering
    \includegraphics[width=0.45\textwidth]{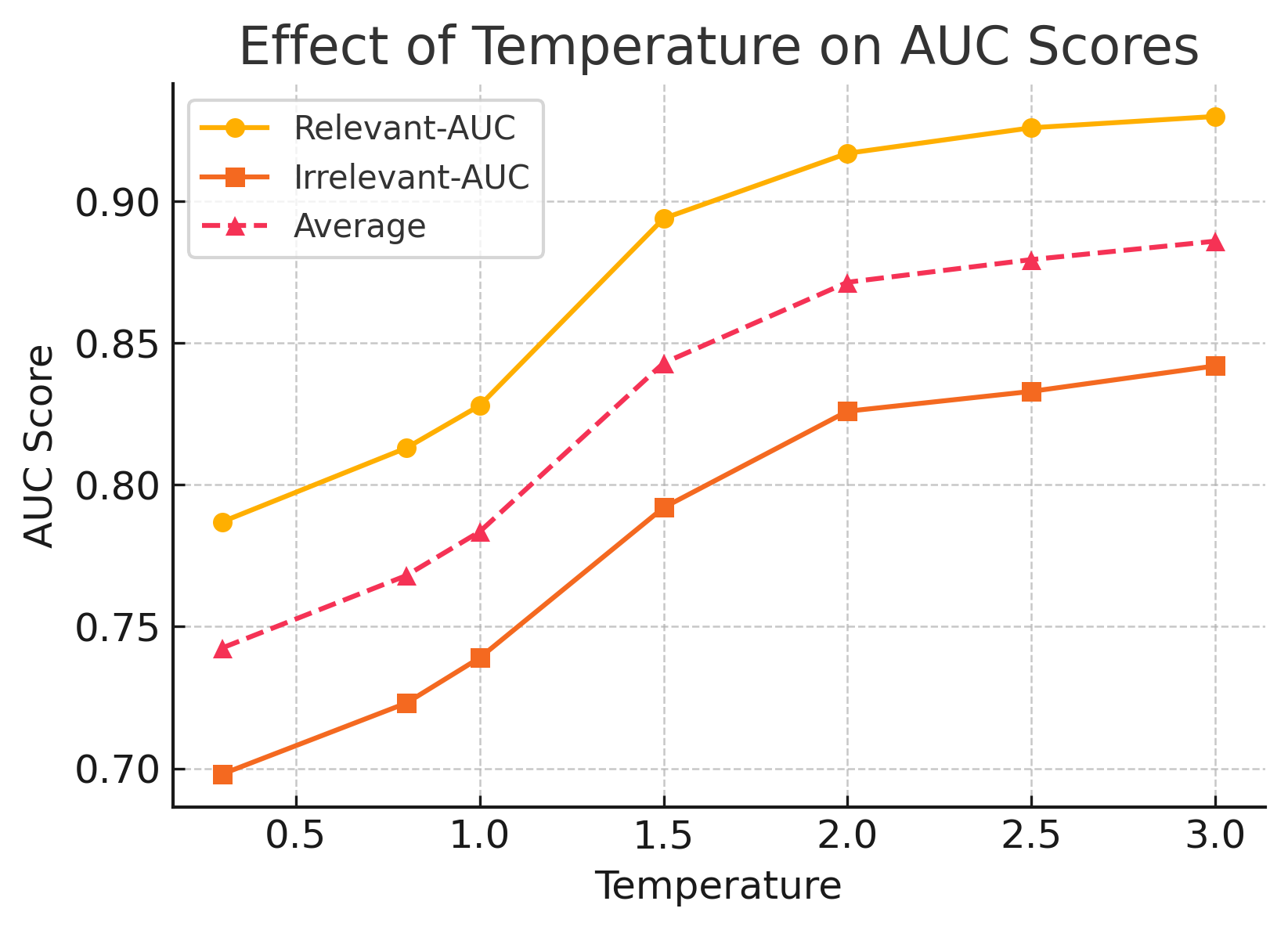}
    \caption{Effect of Temperature on AUC Scores}
    \label{fig:temperature_auc}
\end{figure}

\paragraph{Remarks.}
We set the generation temperature to 3.0 during inference. Unlike general-purpose LLM usage where temperatures below 1.5 are typical, we found that raising the temperature up to approximately 3.0 yielded better results in our fine-tuned scenario. We hypothesize that a lower temperature causes the model to be overly confident in a single token (e.g., probability \(\approx 0.998\)) and thus reduces the meaningful effect of aggregating multiple token probabilities. A summary of these experiments and their outcomes is provided in Figure \ref{fig:temperature_auc}. Mean pooling is used for extracting embedding from $\text{QUPID}_{EMB}$ ( decoder-only model structure).

\subsection{Details of Use Case and Efficiency}
\subsubsection{Filtering low-quality Q-D pairs} \label{subsub: filtering Q-D pairs}
\begin{figure}[ht]
    \centering
    \begin{tikzpicture}[scale=0.8]
        \begin{axis}[
            xlabel={Recall},
            ylabel={Precision},
            title={Precision-Recall Curve},
            legend pos=south west,
            grid=major,
            xmin=0, xmax=1, ymin=0.4, ymax=1
        ]
        
        \addplot[red, solid, mark=square, thick] coordinates {
            (0.0, 1.0) (0.0553, 1.0) (0.1320, 0.9925) (0.3112, 0.9882)
            (0.5721, 0.9701) (0.7434, 0.9213) (0.8518, 0.8348) (0.9340, 0.7567)
            (0.9824, 0.6638) (0.9980, 0.5322) (1.0, 0.4477)
        };
        \addlegendentry{Irrelevant}
        
        \addplot[blue, dashed, mark=o, thick] coordinates {
            (0.0, 0.9945) (0.2891, 0.9945) (0.5967, 0.9766) (0.7566, 0.9339)
            (0.8634, 0.8779) (0.9485, 0.8201) (0.9857, 0.7397) (0.9970, 0.6411)
            (0.9992, 0.5868) (1.0, 0.5664) (1.0, 0.5523)
        };
        \addlegendentry{Relevant}
                
        \draw[dashed, thick, black] (axis cs:0,0.95) -- (axis cs:1,0.95);
        
        \end{axis}
    \end{tikzpicture}
    \caption{PR curve of $\text{QUPID}_{ENSEMBLE}$ model on the \textbf{Web-D} dataset.}
    \label{fig:pr-curve}
\end{figure}
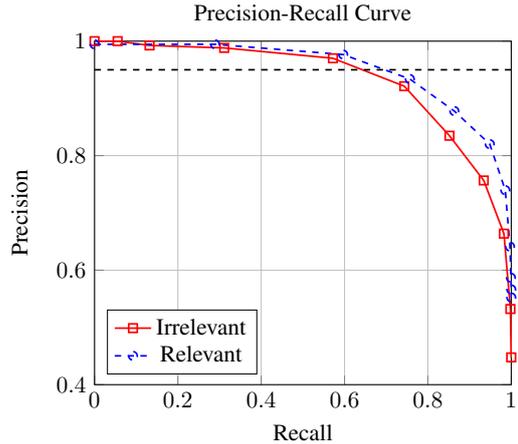

Fundamentally, the trade-off between precision and recall is observed in Figure \ref{fig:pr-curve}. As indicated by the black horizontal dotted line, we can apply thresholding that prioritizes high precision, even at the cost of some coverage (recall). This approach allows us to minimize side effects, such as filtering out high-quality documents, while effectively filtering out low-quality documents with high accuracy (above 0.95).

\subsubsection{Efficiency Compare} \label{appendix:efficiency compare details}
In contrast to the prompting-based approach, the experiment for our model (QUPID) involved the addition of just one special token (\textit{<|task\_prefix|>}) and short system prompt (\textit{query:}, \textit{document:}), alongside the query and document. During the process of fine-tuning on the target task with a vast amount of data, we experimentally confirmed that lengthy prompts were unnecessary. Thus, through the fine-tuning process, we can also observe the advantage of being able to drastically simplify long natural language prompts.

\begin{table}[ht]
\centering
\begin{tabular}{lcc}
\toprule
\textbf{Metric} & \textbf{Search Results} & \textbf{w/ QUPID} \\
\midrule
nDCG@1 & 0.8236 & \textbf{0.8265} \\
nDCG@3 & 0.8511 & \textbf{0.8651} \\
nDCG@5 & 0.8769 & \textbf{0.8938} \\
DCG@1  & 8.7007 & \textbf{9.0549} \\
DCG@3  & 15.6358 & \textbf{16.1056} \\
DCG@5  & 19.3753 & \textbf{19.9544} \\
\bottomrule
\end{tabular}
\caption{Comparison of ranking metrics on 719 queries (each with
an average of 15 documents). This demonstrates that the relevance score can directly serve as a ranking feature, even though no separate ranking loss was introduced and the original architecture and training methodology of QUPID were maintained.}
\label{tab:ranking_performance}
\end{table}

\begin{table*}[t]
    \centering
    \small
    \renewcommand{\arraystretch}{1.2} 
    \setlength{\tabcolsep}{5pt} 
    \begin{tabular}{m{2.5cm} m{4.5cm} | m{2.5cm} m{4.5cm}} 
        \toprule
        \multicolumn{2}{c|}{\textbf{Trump -> \textit{Trump assassination attempt} }} & \multicolumn{2}{c}{\textbf{Biden -> \textit{Biden assassination attempt}}} \\
        \midrule
        \textbf{Original Query} & Trump & \textbf{Original Query} & Biden \\
        \textbf{Auto-Completion} & Trump assassination attempt & \textbf{Auto-Completion} & Biden assassination attempt \\
        \textbf{Relevance Score} &  Rank 1/2/3: 0.804/0.905/0.384 & \textbf{Relevance Score} & Rank 1/2/3:  0.307/0.219/0.135 \\
        \midrule
        \multicolumn{4}{c}{\textbf{Top-3 Retrieved Documents (Title \& Body)}} \\
        \midrule
        \textbf{Rank 1 (Title)} & "Donald Trump rally attack incident" & \textbf{Rank 1 (Title)} & "Trump: 'Assassination attempt due to Biden-Harris rhetoric'" \\
        \textbf{Rank 1 (Body)} & "On July 13, 2024, in Pennsylvania, an assassination attempt was made on Donald Trump during a campaign rally. Security forces responded immediately and neutralized the attacker." & 
        \textbf{Rank 1 (Body)} & "On September 17, 2024, in a Fox News interview, Trump claimed that Biden and Harris were responsible for inciting violence, linking their rhetoric to the assassination attempt." \\
        \midrule
        \textbf{Rank 2 (Title)} & "Trump, second assassination attempt... Secret Service responded in time" & \textbf{Rank 2 (Title)} & "Breaking: Trump, second assassination attempt suspect arrested at golf course" \\
        \textbf{Rank 2 (Body)} & "On September 16, 2024, Secret Service intervened in Florida to prevent another assassination attempt on Trump. The suspect was found carrying a weapon near his residence." & 
        \textbf{Rank 2 (Body)} & "Breaking reports suggest a suspect was arrested at a golf course while attempting another attack on Trump. Officials confirmed White House was briefed immediately." \\
        \midrule
        \textbf{Rank 3 (Title)} & "'Another assassination threat due to Harris' – Trump's claim had no impact on election dynamics" & \textbf{Rank 3 (Title)} & "'Pro-Trump' Musk mocks Biden and Harris over assassination rumors" \\
        \textbf{Rank 3 (Body)} & "Trump suggested that Harris' political influence contributed to threats against him, though polls indicated minimal impact on voter sentiment." & 
        \textbf{Rank 3 (Body)} & "Elon Musk commented on X (formerly Twitter) that no one had ever attempted to assassinate Biden or Harris, sparking controversy online." \\
        \bottomrule
    \end{tabular}
    \caption{Comparison of Top-3 Retrieved Documents (Title \& Body) for Trump and Biden Query Auto-Completion Cases. It can be observed that the relevance score is significantly higher when the auto-completed query corresponds to a realistically searchable query (left) compared to when it does not (right). The text shown in the table was directly used as input, and although the majority of the training data is in Korean, the multilingual capability of the backbone model appears to enable its functionality in English as well.}
    \label{tab:qr_qc_comparison}
\end{table*}

\begin{table*}[ht]
    \centering
    \renewcommand{\arraystretch}{1.2} 
    \begin{tabular}{|p{7.5cm}|p{7.5cm}|}  
        \hline
        \multicolumn{2}{|c|}{Query: \textbf{What companies does Elon Musk own?}} \\
        \hline
        Document \textbf{(Q-D score: 0.812)} & Extracted Snippet \textbf{(Q-S score: 0.284)} \\
        \hline
        \textbf{Elon Musk’s Business Empire: A Look at His Companies}. Elon Musk is one of the most influential entrepreneurs of the 21st century... \textbf{His ventures range from electric vehicles and renewable energy to space exploration and artificial intelligence.} \textbf{He serves as CEO of Tesla, SpaceX, and Neuralink}... 
        & 
        Musk was born in South Africa and later moved to the U.S. to pursue his career. From an early age, he showed a strong interest in technology and entrepreneurship. \\
        \hline
    \end{tabular}
    \caption{Side-by-side comparison of a document and its extracted snippet. The document is relevant to the query (high Q-D relevance score), but the snippet is misleading (low Q-S relevance score). The relevant information to the query is highlighted in bold. We note that the relevance scores shown in this table were obtained by directly feeding the English text as input to the QUPID model.}
    \label{tab:side-by-side-comparison of document and snippet}
\end{table*}


\end{document}